# Applying Nature-Inspired Optimization Algorithms for Selecting Important Timestamps to Reduce Time Series Dimensionality

Muhammad Marwan Muhammad Fuad

Coventry University
School of Computing, Electronics and Mathematics
ad0263@coventry.ac.uk

**Abstract** Time series data account for a major part of data supply available today. Time series mining handles several tasks such as classification, clustering, query-by-content, prediction, and others. Performing data mining tasks on raw time series is inefficient as these data are high-dimensional by nature. Instead, time series are first pre-processed using several techniques before different data mining tasks can be performed on them. In general, there are two main approaches to reduce time series dimensionality; the first is what we call landmark methods. These methods are based on finding characteristic features in the target time series. The second is based on data transformations. These methods transform the time series from the original space into a reduced space, where they can be managed more efficiently. The method we present in this paper applies a third approach, as it projects a time series onto a lower-dimensional space by selecting important points in the time series. The novelty of our method is that these points are not chosen according to a geometric criterion, which is subjective in most cases, but through an optimization process. The other important characteristic of our method is that these important points are selected on a dataset-level and not on a single time series-level. The direct advantage of this strategy is that the distance defined on the low-dimensional space lower bounds the original distance applied to raw data. This enables us to apply the popular GEMINI algorithm. The promising results of our experiments on a wide variety of time series datasets, using different optimizers, and applied to the two major data mining tasks, validate our new method.

**Keywords** Classification. Clustering. Differential evolution. Genetic algorithm· Particle swarm optimization. Time series mining.

## 1 Introduction

A time series is a sequence of measurements recorded over time. Time series data account for a major part of data supply available today. These data appear in many applications ranging from medicine and finance, to sensory data, meteorology and economics.

Time series mining handles several tasks such as classification, clustering, query-by-content, anomaly detection, summarization, segmentation, motif discovery, and others.

The most prominent problem in managing time series is the high dimensionality of these data. Therefore, a high-level representation of time series is a key factor to implementing these tasks efficiently and effectively.

Most time series representation methods follow the *Generic Multimedia Indexing* (GEMINI) framework (Faloutsos et al. 1994) in which all the time series in the dataset are embedded into another lower-dimensional space where, under certain conditions, the different time series mining tasks can be handled more efficiently. More formally, let $n$ be the dimensionality of the raw data, and $N$ be the dimensionality of the transformed space. $N$ should satisfy: $N < n$ (ideally, $N \ll n$). $d^N$ is said to be *lower-bounding* of $d^n$ if $d^N(\bar{S},\bar{T}) \leq d^n(S,T)$, where $\bar{S}$, $\bar{T}$ are the projections of the time series $S$, $T$, respectively, on the transformed space.

There have been several alternatives to applying different time series data mining tasks directly to raw data. These alternatives can be divided into two main categories: *landmark methods* and *data transformation-based methods*.

In a previous work (Muhammad Fuad 2016) we presented the outline of a third alternative that addresses the above problem differently. This method transforms the data into a low-dimensional space by choosing important points, but unlike landmark methods, the points in our method are not selected based on geometric properties, but on a classification-error basis, where this classification is performed in the framework of an optimization process using differential evolution as optimizer. The direct result of this method, and because all the points in the transformed space are points in the original data, is that the distance between any two time series in the transformed space is a lower bound of the two corresponding time series in the original space. Consequently, the GEMINI algorithm can be applied without any additional procedures.

In this text we extend our method presented in (Muhammad Fuad 2016) by using different optimizers applied to several time series mining tasks. The outcome of the optimization process is the timestamps that yield the best results in a time series mining task. The optimizers we use in this paper are differential evolution, genetic algorithm, and particle swarm optimization. The time series mining tasks we apply our method in this work to are classification and clustering. In addition, and this is a particular novelty of this paper, we show how our method can be built to optimize several time series mining tasks simultaneously. The optimizer we use in the multi-objective optimization process is the popular non-dominated sorting genetic algorithm-II (NSGA-II). All the variations of our method we present in this work are compared against traditional time series representation methods on a wide variety of datasets. The results show a clear superiority of our method over traditional time series representation methods.

This paper is organized as follows; Section 1 is a background section. Section 2 introduces traditional time series representation methods. Our new method is motivated and presented in Section 3. In Section 4 we conduct experiments that validate our method. Section 5 is a concluding section, where we discuss the outcome of our experiments and suggest some directions for future research.

## 2 A brief introduction to data mining

Data mining is one of the branches of computer science that witnessed substantial progress in the last years. Data mining encompasses several tasks the main of which are (Bramer 2007), (Gorunescu 2006), (Larose 2005), (Mörchen 2006):

- **Data Pre-processing**: Most raw data are unprepared, noisy, or incomplete. For this reason, a preparation stage is required before handling data. This stage may include different processes such as data cleansing, normalization, handling outliers, completion of missing values, and deciding which attributes to keep and which ones to discard.

- **Prediction:** This task can be viewed as forecasting the future state of a phenomenon given its current state. Prediction is similar to estimation, except that it concerns values that are beyond the range of already observed data.

- **Query-by-content:** In this task the algorithm searches for all the objects in the database that are similar to a given pattern.

- **Classification:** Classification is the task of assigning items to predefined classes. Classification is one of the main tasks of data mining and it is particularly relevant to the experimental section of this paper. There are a number of classification models, the most popular of which is $k$-*nearest-neighbor* ($k$NN). In this model the object is classified based on the $k$ closest objects in its neighborhood. Performance of classification algorithms can be evaluated using different methods. One of the widely used ones is *leave-one-out cross-validation* (LOOCV) - also known by *N-fold cross-validation*, or *jack-knifing*, where the dataset is divided into as many parts as there are instances, each instance effectively forming a test set of one. N classifiers are generated, each from $N-1$ instances, and each is used to classify a single test instance. The classification error is then the total number of misclassified instances divided by the total number of instances (Bramer 2007).

- **Clustering:** It is the task of partitioning the data objects into groups, called *clusters*, so that the objects within a cluster are similar to one another and dissimilar to the objects in other clusters (Han et al. 2011). Clustering differs from classification in that there is no target variable for clustering. Instead, clustering algorithms attempt to segment the entire data set into relatively homogeneous subgroups or clusters (Larose 2005).

There are several basic clustering methods such as: *Partitioning Methods*, *Hierarchical Methods*, *Density-Based Methods*, and *Grid-Based Methods*.

$k$-*means* is one of the most widely used and studied clustering formulations (Kanungo et al. 2002). $k$-means is a centroid-based partitioning technique which uses the *centroid* (also called *center*) of a cluster; $c_i$, to represent that cluster. Conceptually, the centroid of a cluster is its center point. The centroid is defined as the mean of the objects assigned to the cluster.

In $k$-means clustering we have a set of $n$ data points in $d$-dimensional space $R^d$ and an integer $k$, the problem is to determine a set of $k$ points in $R^d$, the centroids, so as to minimize the mean distance from each data point to its nearest center (Kanungo et al. 2002).

More formally, the $k$-means clustering error can be measured by:

$$E = \sum_{i=1}^{k} \sum_{j=1}^{n_j} d(u_{ij}, c_i) \qquad (1)$$

The number of clusters is decided by the user, application-dependent, or given by some cluster validity measure.

## 3 Time series mining

Many data mining tasks involve temporal aspects. The most common example of these temporal data is time series. The main characteristic of time series data is their high-dimensionality. Representation methods are widely used to manage the high-dimensionality of time series (hence they are also called *dimensionality reduction techniques*). In addition to reducing dimensionality, representation methods also help emphasize certain features of the data, in addition to removing noise and speeding up the different time series mining tasks (Mörchen 2006). They also reduce storage space.

Time series representation methods can be divided into two main categories: the first is what we call landmark methods (Hetland 2003). These methods are based on finding characteristic features in the target time series. In (Perng et al. 2000) the authors present a model which allows any point of great importance to be identified as a landmark. The gist of the landmark model is to use landmarks instead of raw data. For instance, first-order landmarks are extreme points, second-order landmarks are inflection points, and so on. The model takes into account that local extreme points are not as important as global extreme ones. In Figure 1 we illustrate the landmark model of (Perng et al. 2000). The data presented in this figure is taken from dataset (SonyAIBORobotSurfaceII) available at (Chen et al. 2015) – we took only the first 16 data points of each time series. The time series are shifted vertically in the figure for clarity. For each time series the landmark model selects the important points of each time series according to a geometric criterion as mentioned above (we took four points from each time series - shown in the figure as red bold dots,). The main point to mention here is that in the landmark model the important points of each time series are selected independently of the selection of important points of the other time series in the dataset, as we can see from the figure. This, as we will see later, is one of the main differences between the landmark model and our method, which selects the important points on a dataset-level.

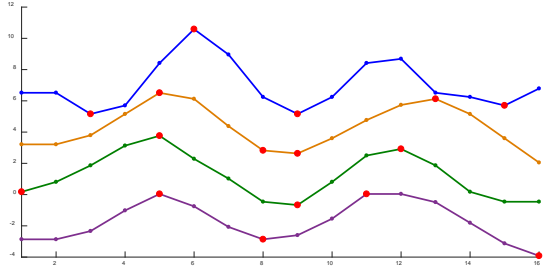

**Fig. 1.** The landmark model of four time seroes indicating-in bold red dots – the landmark points of each time series.

Landmark methods offer a low-dimension representation that is usually invariant to some transformations such as time-warping, shifting, uniform amplitude scaling, non-uniform amplitude scaling, and uniform time scaling (Wang and Megalooikonomou 2008). The main drawback of the landmark methods, in our opinion, is that the choice of important points is subjective and generic as it applies the same criteria to all the data without taking into account that what is considered an important landmark point for a certain dataset is not necessarily important for another. An extreme point, for instance, may be considered a landmark point because it reflects a change in time series behavior for a certain dataset, whereas it may simply be the result of noise for another dataset.

The second category of representation methods is based on data transformations. These methods apply the GEMINI algorithm which we presented in the previous section.

Several representation methods have been proposed in the literature. Examples of the most common ones include *Discrete Fourier Transform* (DFT) (Agrawal et al. 1993) and (Agrawal et al. 1995), *Discrete Wavelet Transform* (DWT) (Chan and Fu 1999), *Singular Value Decomposition* (SVD) (Korn et al. 1997), *Adaptive Piecewise Constant Approximation* (APCA) (Keogh et al. 2001), *Piecewise Linear Approximation* (PLA) (Morinaka et al. 2001), and *Chebyshev Polynomials* (CP) (Cai and Ng 2004).

In the following we present a description of two dimensionality reduction techniques which are related to the experimental section of this paper:

a- **The Piecewise Aggregate Approximation (PAA):** PAA (Keogh et al. 2000) (Yi and Faloutsos 2000) divides a time series $S$ of $n$-dimensions into equal-sized segments and maps each one to a point of a lower $N$-dimensional space, where each point in the reduced space is the mean value of the data points falling within that segment. The similarity measure given in the following equation:

$$d^N(S,T) = \sqrt{\frac{n}{N}} \sqrt{\sum_{i=1}^{N}(\overline{s_i} - \overline{t_i})^2} \qquad (2)$$

is defined on the $N$-dimensional space. This similarity measure is a lower bound of the Euclidean distance defined on the original $n$-dimensional space.

It is worth mentioning that the authors of PAA use a compression ratio of 1:4 (i.e. every 4 points in the original time series are represented by 1 point in the reduced space) when applying PAA. This remark is related to the experimental part of our paper.

b- **The Symbolic Aggregate approximation (SAX):** SAX (Lin et al. 2003) is one of the most powerful symbolic representation methods of time series. The main advantage of SAX is that the similarity measure it applies, called MINDIST, uses statistical lookup tables, which makes it easy to compute. SAX is applied as follows:

1. The time series are normalized.
2. The dimensionality of the time series is reduced using PAA.
3. The PAA representation of the time series is discretized.

## 4 Time series dimensionality reduction by optimized selection of timestamps

### 4.1 Motivation and Principle

The main drawback of traditional time series representation methods that we mentioned in the previous section is that the resulting lower-dimensional representation of the time series may have a strong smoothing effect that important local information can be lost. We give the following example to illustrate this point: given the four time series $S_1 = [+1, -4, +11, 0, +3, -9, +4, 0]$, $S_2 = [-1, +10, -5, +4, -5, +5, -3, +1]$, $S_3 = [+2, -1, +3, 0, -5, -4, -2, -1]$, and $S_4 = [-8, +10, +17, -15, -18, +9, +4, -7]$. Let us use PAA (with a compression ratio of 1:4), which we introduced in the previous section, to reduce the dimensionality of these time series. When comparing $S_1$ to $S_2$ we see that their PAA representation is the same, which is $PAA = [+2, -0.5]$ (Figure 2a), although, as we can see, $S_1$ and $S_2$ are not similar. This is also the case when comparing $S_3$ to $S_4$ (Figure 2b), which have the same PAA representation: $[+1, -3]$ although their shapes are very different. This problem may also appear with other time series representation methods.

Our new approach of time series representation remedies the drawbacks appearing in the landmark-based methods and data

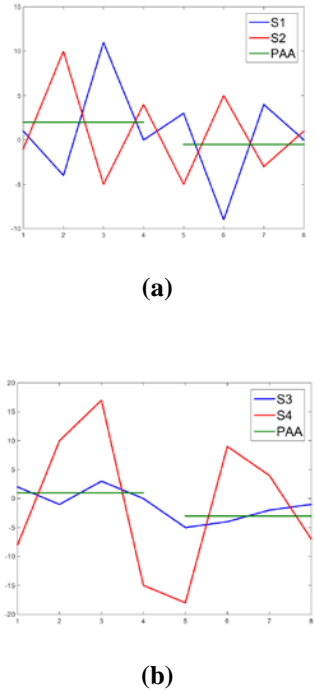

**Fig. 2.** PAA representation.

transformation-based methods. As in the case with landmark-based methods, our new method also keeps only the important points that we project onto a low-dimension space, but there are two main differences between our method and landmark-based methods:

1- The choice of important points in our method is objective and is not based on *a priori* knowledge of the data.

2- The choice of important points is not based on individual time series, i.e. it does not select important points for each time series individually, but it selects the important points on a dataset level. In other words, we select the indices of important timestamps for the whole dataset and we reduce the dimensionality by keeping, for each time series, the data points of that time series that correspond to those timestamps. This dataset-based selection strategy of important timestamps is one of the two main features of our method and it has several advantages:

   - Since each time series in the dataset will be projected on the low-dimension space by keeping the points of that time series that correspond to the indices of the important timestamps of the dataset, and given that the distance defined on the original space is the Euclidean distance, one can easily prove that the distance defined on the low-dimension space is a lower bound of the original distance because it is simply a sub-sum of the original distance. This important condition is not maintained in traditional landmark methods.

   - Selecting the important timestamps on a dataset level will have a positive smoothing effect, as this will eliminate noise. This smoothing effect is different from the one we discussed earlier in this section as it will still keep the general pattern of the time series.

   - This choice is in fact an intuitive one. Time series datasets record how a certain observation is measured at different timestamps, where each time series records how a phenomenon is observed on a specific entity. For example, in gene expression data each time series in the dataset may express how a specific process progresses over time for a certain organism or individual, so the timestamp represents an event, and the observer is interested in examining the value of each time series in the dataset at a certain "important" event.

   - Whereas other dimensionality reduction techniques apply standard compressions ratios (1:4, 1:6, etc.) our method applies a fully customized dimensionality reduction defined by the number of kept points determined by the user. As a matter of fact, and although we use certain compression ratios in the experimental section of this work, we only do this to compare the performance of our method to that of other methods. But the concept of a "compression ratio" does not make sense in our method as it is based on the number of timestamps which is independent of the original dimensionality of the raw data.

The second main feature of our method is that the important timestamps are selected through an optimization process. But before we introduce our method, we first start by giving a brief introduction to optimization.

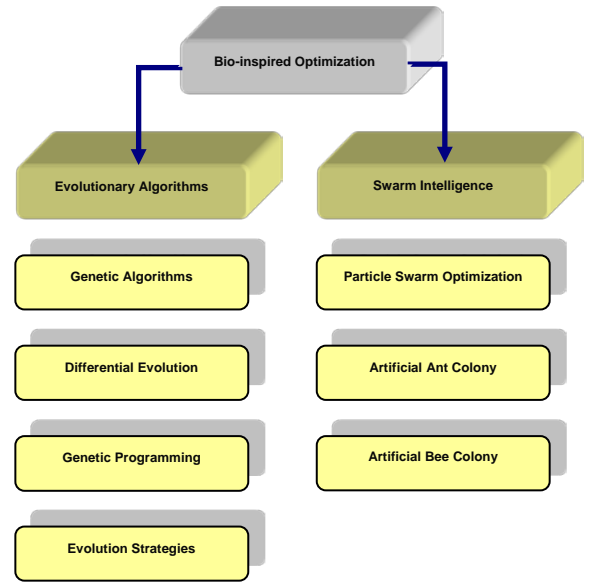

**Fig. 3.** The two main families of natura-inspired optimzation algorithms.

Optimization is the process of finding the best solution to a problem under certain constraints. More formally, optimization can be defined as follow; given a function $f$ of $nbp$ parameters: $f: U \subseteq \mathbb{R}^{nbp} \to \mathbb{R}$ which we call the *fitness function*, find the solution $\overrightarrow{X^*} = [x_1^*, x_2^*, \ldots, x_{nbp}^*]$ which satisfies $f(\overrightarrow{X^*}) \leq f(\vec{X})\ \forall \vec{X} \in U$.

Optimization algorithms can be classified in several ways, one of which is whether they are *single solution –based* algorithms; these use one solution and modify it to get the best solution. The other category is *population-based* algorithms; these use several solutions which exchange information to find the best solution.

Optimization problems can be handled using deterministic algorithms or probabilistic ones. *Metaheuristics* are probabilistic optimization algorithms applicable to a large variety of optimization problems. Metaheuristics are usually applied when the search space is very large, or when the number of parameters of the optimization problem is very high, or when the relationship between the fitness function and the parameters is not clear. Many of these metaheuristics are inspired by natural processes, natural phenomena, or by the collective intelligence of natural agents, hence the term *nature-inspired* or *bio-inspired* optimization algorithms.

Nature-inspired optimization can be classified into two main families; the first is *Evolutionary Algorithms* (EA). This family is probably the largest family of nature-inspired algorithms. EA are population-based algorithms that use the mechanisms of Darwinian evolution such as selection, crossover and mutation. Of this family we mention: *Genetic Algorithms* (GA), *Genetic Programming* (GP), *Evolution Strategies* (ES), and *Differential Evolution* (DE). The other family is *Swarm Intelligence* (SI). This family uses algorithms which simulate the behavior of an intelligent biological system. Of this family we mention: *Particle Swarm Intelligence* (PSO), *Ant Colony Optimization* (ACO), and *Artificial Bee Colony* (ABC). Fig. 3 shows the main nature-inspired optimization algorithms.

In the following we present a brief description of the optimization algorithms we apply in this paper:

### 4.2 Genetic Algorithm (GA)

GA is one of the most prominent global optimization algorithms. Classic GA starts by randomly generating a population of chromosomes that represent possible solutions to the problem at hand. Each chromosome is a vector whose length is equal to the number of parameters, denoted by $nbp$. The fitness function of each chromosome is evaluated in order to determine the chromosomes that are fit enough to survive and possibly mate. In the *selection* step a percentage $sRate$ of chromosomes is selected for mating. *Crossover* is the next step in which the offspring of two parents are produced to enrich the population with fitter chromosomes. *Mutation*, which is a random alteration of a certain percentage $mRate$ of chromosomes, enables GA to explore the search space. In the next generation the fitness function of the offspring is calculated and the above steps repeat for a number of generations $nGen$.

### 4.3 Differential Evolution (DE)

DE is one of the most powerful evolutionary optimization algorithms. DE starts with a population of $popSize$ vectors each of which is of $nbp$ dimensions. Next, for each individual $\vec{T}_i$ (called the *target vector*) of the population three mutually distinct individuals $\vec{V}_{r1}$, $\vec{V}_{r2}$, $\vec{V}_{r3}$ and different from $\vec{T}_i$ are chosen randomly from the population. The *donor vector* $\vec{D}$ is formed as a weighted difference of two of $\vec{V}_{r1}, \vec{V}_{r2}, \vec{V}_{r3}$ added to the third, i.e. $\vec{D} = \vec{V}_{r1} + F(\vec{V}_{r2} - \vec{V}_{r3})$. $F$ is called the *mutation factor*. The *trial vector* $\vec{R}$ is formed from elements of the target vector $\vec{T}_i$ and elements of the donor vector $\vec{D}$ according to different schemes. In this paper we choose the crossover scheme presented in (Feoktistov 2006). In this scheme an integer $Rnd$ is chosen randomly among the dimensions $[1, nbp]$. Then the trial vector $\vec{R}$ is formed as follows:

$$t_i = \begin{cases} t_{i,r1} + F(t_{i,r2} - t_{i,r3}) \\ \qquad if\ (rand_{i,j}[0,1[ < C_r) \vee (Rnd = i) \\ t_{i,j} \\ \qquad otherwise \end{cases} \quad (3)$$

where $i = 1, \ldots, nbp$. $C_r$ is the *crossover constant*.

In the next step DE selects which of the trial vector and the target vector will survive in the next generation and which will die out. This selection is based on which of $\vec{T}_i$ and $\vec{R}$ yields a better value of the fitness function. The algorithm iterates for a number of generations $nGen$.

### 4.4 Particle Swarm Optimization (PSO)

PSO is inspired by the social behavior of some animals, such as bird flocking or fish schooling (Haupt and Haupt 2004). In PSO individuals, called *particles*, follow three rules a) *Separation:* each particle avoids getting too close to its neighbors. b) *Alignment:* each particle steers towards the general heading of its neighbors, and c) *Cohesion:* each particle moves towards the average position of its neighbors.

PSO starts by initializing a swarm of $popSize$ particles at random positions $\vec{X}_i^0$ and velocities $\vec{V}_i^0$ where $i \in \{1, \ldots, popSize\}$. In the next step the fitness function of each position, and for each iteration, is evaluated. The positions $\vec{X}_i^{k+1}$ and velocities $\vec{V}_i^{k+1}$ are updated at time step $(k+1)$ according to the following formulae:

$$\vec{V}_i^{k+1} = \omega.\vec{V}_i^k + \varphi_G(\vec{G}^k - \vec{X}_i^k) + \varphi_L(\vec{L}_i^k - \vec{X}_i^k) \quad (4)$$

$$\vec{X}_i^{k+1} = \vec{X}_i^k + \vec{V}_i^k \quad (5)$$

where $\varphi_G = r_G.a_G$, $\varphi_L = r_L.a_L$, $r_G, r_L \to U(0,1)$, $\omega, a_L, a_G \in \mathbb{R}$. $\vec{L}_i^k$ is the best position found by particle $i$, $\vec{G}^k$ is the global best position found by the whole swarm, $\omega$ is called the *inertia*, $a_L$ is called the *local acceleration*, and $a_G$ is called the *global* *acceleration*. The algorithm continues for a number of iterations $nGen$.

**4.5 Non-dominated Sorting Genetic Algorithm II (NSGA-II)**

Although single-objective optimization problems are widely-encountered, many practical optimization problems have to satisfy several criteria that are conflicting in many cases. This class of optimization problems is called *Multi-objective Optimization* (MOO). An $m$-dimensional MOO problem can be formulated as follows:

$$min\{f_1(\vec{X}), f_2(\vec{X}), ..., f_m(\vec{X})\}$$

Where $\vec{X} \in \mathbb{R}^{nbp}$

The optimal solution for MOO is not a single solution as for single-objective optimization problems, but a set of solutions defined as *Pareto optimal solutions* (El-Ghazali T. 2009), also called a *non-dominated* solution. A solution is Pareto optimal if it is not possible to improve a given objective without deteriorating at least another objective.

The non-dominated sorting genetic algorithm (NSGA) (Srinivas and Deb 1995) is one of the most popular algorithms to solve MOO. In NSGA, all non-dominated individuals are classified into one category, with a dummy fitness value proportional to the population size. This group is then removed and the remaining population is reclassified. The process is repeated until all the individuals in the entire population are classified. A stochastic remainder proportionate selection is used (Maulik et al. 2011).

NSGA, however, has been criticized for its high computational cost, its lack of elitism, and for its need to specify the sharing parameter. For these reasons, NSGA-II was proposed in (Deb et al. 2002). NSGA-II can be summarized as follows (Ma at al. 2007): a random parent population is initialized. The population is sorted based on non-domination in two fronts, the first front being completely a non-dominant set in the current population and the second being dominated by the individuals in the first front only. Each solution is assigned a rank equal to its non-domination level based on the front it belongs to. Individuals in the first front are assigned a fitness value of 1 and individuals in the second are assigned a fitness value of 2 and so on. The authors of NSGA-II introduce a new parameter called the *crowding distance*. This parameter measures how close every individual is to its neighbors. The crowding distance is calculated for each individual of the population. Parents are selected from the population by using a binary tournament selection based on the rank and the crowding distance. An individual is selected if its rank is less than that of the other or if its crowding distance is greater than that of the other. The selected population generates offspring from crossover and mutation operators. The population with the current population and current offspring is sorted again based on non-domination and only the best $popSize$ individuals are selected, where $popSize$ is the population size. The selection is based on the rank and on the crowding distance on the last front.

The process repeats to generate the subsequent generations.

**4.6 Our Proposed Method**

In the following we present a detailed description of our method as presented in Figure 4. The algorithm designer selects the optimization algorithm to be used for the optimization process and the corresponding control parameters of that optimizer. We tested three optimizers in the experimental section of this paper.

| **Algorithm:** Optimization Algorithm for Time Series Dimensionality Reduction |
|---|
| **Input:** The selected optimizer (GA,DE,PSO, NSGA-II). |
| **Input:** The selected fitness function (1NN classification error , k-means clustering quality). |
| **Input:** Control parameters (popSize, nGen, Nbp, etc). |
| **Data:** The training set of the dataset in question. |
| **Output:** The timestamps of that dataset which yield the optimal value of the fitness function on the training set. |
| 1: Randomly initialize popSize chromosomes of length nbp. Each chromosome is a vector whose components are integers, between 1 and the dimension of the time series, in ascending order. The chromosome corresponds to timestamps of the dataset. |
| 2: For each chromosome, reduce the dimensionality of the dataset by keeping , from each time series, only the data points that correspond to the timestamps indicated by that chromosome. |
| 3: Calculate the fitness function of each chromosome by performing the selected time series mining task (1NN classification error , k-means clustering quality) on the timestamps that correspond to each chromosome. |
| 4: Apply the optimization operations related to the selected optimizer (ranking, selection, mutation, crossover, etc) on the chromosomes. |
| 5: Repeat steps 2-4 for a number of generations nGen |

**Fig. 4.** Nature-inspired optimization algorithm for reducing time series dimensionality.

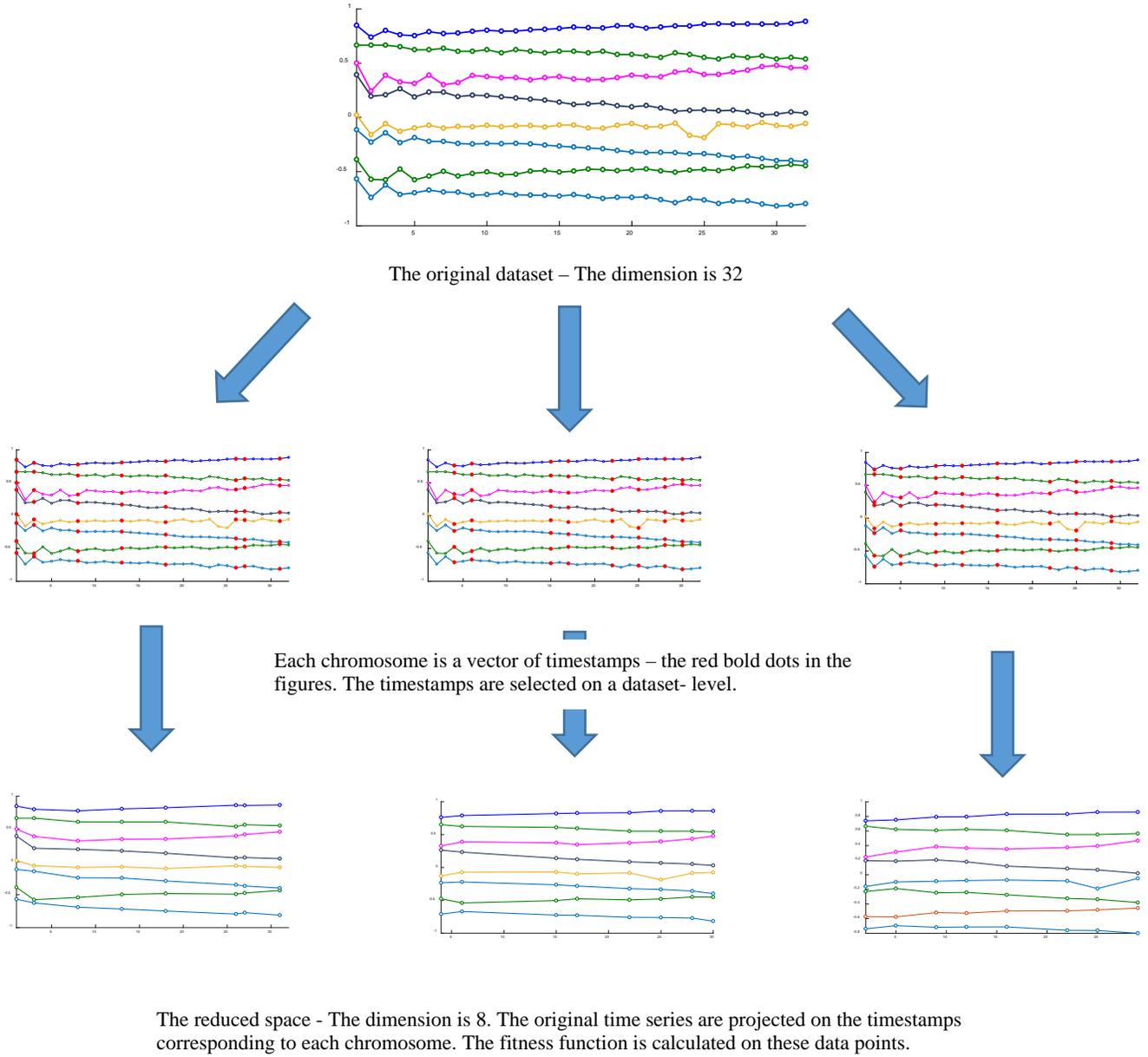

**Fig. 5.** Illustration of three chromsomes in our method. The dimension of the reduced spaces in this figure is 8, which is *nbp* in our algorithm.

However, other appropriate optimizers can be used.

The algorithm designer also selects the time series mining task that will be used as a fitness function for the optimization process. In other words, this fitness function will serve as a criterion for selecting the important timestamps of the dataset in question. This can also be handled as a multi-objective optimization problem where more than one criterion is used to select the important timestamps. The algorithm designer also selects the desired value for $nbp$, which corresponds to the reduced dimensionality of the time series.

The optimization process, whose purpose is to find the important timestamps of that dataset, is applied to the training set of the dataset. The outcome of the optimization process – the optimal timestamps – is then applied to the corresponding testing set for validation.

The optimization process starts by randomly generating a population of chromosomes whose size is $popSize$. Each chromosome is a vector of timestamps of that dataset. The length of the chromosome is $nbp$. Its components are integers of ascending order chosen randomly between 1 and the

Table 1 The control parameters of GA, DE, and PSO

| Symbol | Name | Related Algorithm | Value |
|---|---|---|---|
| $popSize$ | Population size | GA, DE, PSO | 16 |
| $nGen$ | Number of generations | GA, DE, PSO | 100 |
| $nbp$ | Number of parameters | GA, DE, PSO | varies |
| $sRate$ | Crossover rate | GA | 0.5 |
| $mRate$ | Mutation rate | GA | 0.2 |
| $F$ | Differentiation constant | DE | 0.9 |
| $C_r$ | Crossover constant | DE | 0.5 |
| $a_L$ | Local acceleration | PSO | 2 |
| $a_G$ | Global acceleration | PSO | 2 |
| $\omega$ | Inertia | PSO | $\omega = (nGen - cGen^*)/(nGen)$ |

$cGen^*$: current generation

dimension of the time series dataset. The chromosome represents a potential solution of the optimization problem, i.e. the timestamps that yield the optimal value of the fitness function. In order to calculate the fitness function (1NN classification error, $k$-means clustering quality) that corresponds to that chromosome, each time series in the dataset is projected onto the timestamps defined by that chromosome. In other words, for each time series, we discard all the data points except those that correspond to the timestamps indicated by that chromosome. We apply the selected time series mining task on this reduced dataset to calculate the value of the fitness function that corresponds to that chromosome. We proceed in the same manner for each chromosome to calculate the fitness function that corresponds to each chromosome.

Depending on the optimizer we selected, we proceed by applying the optimization operations that correspond to that optimizer (mutation, selection, crossover, etc). By continuing to do so we obtain fitter chromosomes, which correspond to timestamps that yield better results in the classification or clustering task. These timestamps are the important timestamps that establish the reduced space of the time series for that dataset.

Table 2 The 1NN classification error of GA, DE, PSO, PAA, and SAX

| Datasets | 1NN Classification Error | | | | | | | | | | | | | |
|---|---|---|---|---|---|---|---|---|---|---|---|---|---|---|
| | GA | | | | DE | | | | PSO | | | | PAA | SAX |
| | $\rho$=1:4 | $\rho$=1:8 | $\rho$=1:12 | $\rho$=1:16 | $\rho$=1:4 | $\rho$=1:8 | $\rho$=1:12 | $\rho$=1:16 | $\rho$=1:4 | $\rho$=1:8 | $\rho$=1:12 | $\rho$=1:16 | $\rho$=1:4 | $\rho$=1:4 |
| Gun_Point | 0.087 | 0.120 | **0.060** | 0.073 | 0.087 | 0.093 | 0.073 | 0.080 | 0.100 | 0.100 | **0.060** | 0.107 | 0.093 | 0.147 |
| OSULeaf | 0.467 | 0.492 | 0.479 | **0.455** | 0.483 | 0.463 | 0.488 | 0.492 | 0.483 | 0.496 | 0.500 | 0.500 | 0.488 | 0.475 |
| Trace | 0.230 | 0.180 | 0.180 | 0.170 | **0.120** | 0.130 | **0.120** | **0.120** | 0.190 | **0.120** | 0.130 | 0.140 | 0.250 | 0.370 |
| FaceFour | 0.205 | 0.170 | 0.182 | 0.148 | 0.148 | 0.216 | 0.159 | **0.114** | 0.182 | 0.148 | 0.193 | 0.170 | 0.205 | 0.227 |
| ECG200 | 0.100 | 0.120 | 0.160 | 0.130 | 0.120 | 0.130 | 0.130 | 0.150 | 0.140 | **0.080** | 0.120 | 0.120 | 0.130 | 0.120 |
| Adiac | 0.379 | 0.361 | 0.402 | 0.402 | **0.340** | 0.348 | 0.389 | 0.381 | 0.361 | 0.361 | 0.371 | 0.343 | 0.404 | 0.867 |
| FISH | 0.211 | **0.189** | 0.211 | 0.223 | 0.194 | 0.194 | 0.200 | 0.217 | 0.217 | **0.189** | 0.217 | 0.206 | 0.217 | 0.263 |
| Plane | 0.038 | **0.010** | **0.010** | 0.019 | 0.029 | 0.029 | 0.019 | 0.038 | 0.029 | 0.019 | 0.029 | 0.019 | 0.038 | 0.029 |
| Car | 0.250 | 0.233 | **0.200** | 0.267 | 0.250 | 0.250 | 0.233 | 0.250 | 0.250 | 0.233 | **0.200** | 0.233 | 0.267 | 0.267 |
| Beef | 0.333 | 0.300 | 0.233 | 0.300 | 0.300 | 0.333 | 0.267 | 0.267 | 0.300 | 0.233 | **0.167** | **0.167** | 0.333 | 0.433 |
| Coffee | **0.000** | **0.000** | **0.000** | **0.000** | **0.000** | **0.000** | **0.000** | **0.000** | **0.000** | **0.000** | **0.000** | 0.036 | **0.000** | 0.286 |
| CinC_ECG_torso | 0.095 | 0.109 | 0.091 | 0.085 | 0.094 | 0.104 | 0.087 | 0.120 | 0.100 | 0.087 | 0.087 | 0.100 | 0.104 | **0.073** |
| ChlorineConcentration | 0.322 | 0.326 | 0.293 | 0.363 | 0.282 | 0.230 | 0.203 | **0.188** | 0.284 | 0.289 | 0.239 | 0.233 | 0.390 | 0.582 |
| DiatomSizeReduction | 0.059 | 0.049 | **0.042** | 0.056 | 0.056 | 0.049 | 0.046 | 0.052 | 0.052 | 0.052 | 0.046 | 0.046 | 0.065 | 0.082 |
| ECGFiveDays | 0.139 | 0.124 | 0.072 | 0.130 | 0.095 | **0.034** | 0.055 | 0.091 | 0.074 | 0.057 | 0.151 | 0.153 | 0.146 | 0.150 |
| Haptics | 0.610 | 0.610 | 0.600 | 0.607 | 0.607 | **0.578** | 0.581 | 0.584 | 0.588 | 0.581 | 0.581 | **0.578** | 0.643 | 0.643 |
| ItalyPowerDemand | 0.045 | **0.031** | 0.034 | 0.040 | **0.031** | 0.040 | 0.033 | 0.040 | 0.068 | 0.095 | 0.033 | 0.040 | 0.068 | 0.192 |
| MALLAT | 0.081 | 0.086 | 0.077 | 0.083 | 0.089 | 0.084 | 0.096 | 0.094 | 0.086 | 0.091 | **0.068** | 0.078 | 0.089 | 0.143 |
| MoteStrain | **0.129** | 0.132 | 0.181 | 0.181 | 0.139 | 0.169 | 0.164 | 0.161 | 0.141 | 0.171 | 0.158 | 0.148 | 0.190 | 0.212 |
| TwoLeadECG | 0.244 | 0.246 | 0.277 | 0.277 | 0.251 | 0.242 | 0.279 | 0.277 | **0.216** | 0.226 | 0.225 | 0.281 | 0.283 | 0.309 |
| ArrowHead | 0.189 | 0.223 | 0.183 | 0.194 | 0.177 | 0.183 | 0.171 | **0.137** | 0.177 | 0.183 | 0.217 | 0.166 | 0.206 | 0.246 |
| BirdChicken | 0.450 | 0.350 | 0.400 | 0.300 | 0.400 | 0.400 | 0.400 | 0.300 | 0.400 | 0.400 | 0.400 | **0.200** | 0.450 | 0.350 |
| Herring | 0.422 | 0.453 | 0.437 | **0.391** | 0.437 | 0.437 | 0.453 | 0.516 | 0.422 | 0.406 | 0.437 | 0.437 | 0.484 | 0.406 |
| ProximalPhalanxTW | 0.267 | **0.252** | 0.270 | 0.257 | **0.252** | 0.275 | 0.300 | 0.275 | 0.280 | 0.265 | 0.275 | 0.292 | 0.280 | 0.370 |
| DistalPhalanxOutlineAgeGroup | 0.220 | **0.182** | 0.250 | 0.220 | 0.202 | 0.205 | 0.217 | 0.222 | 0.220 | 0.205 | 0.215 | 0.220 | 0.235 | 0.267 |
| Earthquakes | 0.233 | 0.268 | 0.265 | 0.310 | 0.208 | 0.202 | 0.217 | 0.245 | **0.180** | 0.233 | 0.220 | 0.242 | 0.311 | **0.180** |
| MiddlePhalanxOutlineAgeGroup | 0.250 | 0.275 | 0.265 | 0.285 | 0.250 | 0.255 | 0.245 | 0.240 | **0.232** | 0.257 | 0.250 | 0.235 | 0.277 | 0.242 |
| ShapeletSim | 0.483 | 0.456 | 0.489 | 0.494 | 0.444 | 0.489 | 0.500 | 0.461 | **0.417** | 0.467 | 0.483 | 0.456 | 0.472 | 0.428 |
| Wine | 0.352 | 0.352 | **0.259** | 0.333 | 0.352 | 0.278 | 0.333 | 0.352 | 0.370 | 0.407 | **0.259** | 0.352 | 0.370 | 0.500 |
| Strawberry | 0.051 | 0.046 | 0.054 | 0.057 | 0.046 | 0.052 | 0.057 | 0.046 | 0.047 | **0.041** | 0.057 | 0.054 | 0.062 | 0.328 |

Table 3 The training time of the classification task for GA, DE, and PSO

| Datasets | GA | DE | PSO |
|---|---|---|---|
| Gun_Point | 00m 07s | 00m 03s | 00m 14s |
| OSULeaf | 04m 03s | 04m 12s | 03m 50s |
| Trace | 01m 00s | 01m 02s | 00m 59s |
| FaceFour | 00m 06s | 00m 05s | 00m 05s |
| ECG200 | 00m 55s | 00m 58s | 00m 54s |
| Adiac | 13m 36s | 13m 28s | 13m 14s |
| FISH | 03m 07s | 03m 04s | 02m 57s |
| Plane | 01m 01s | 01m 03s | 00m 59s |
| Car | 00m 29s | 00m 26s | 00m 25s |
| Beef | 00m 10s | 00m 07s | 00m 07s |
| Coffee | 00m 00s | 00m 00s | 00m 06s |
| CinC_ECG_torso | 00m 31s | 00m 17s | 00m 17s |
| ChlorineConcentration | 19m 16s | 19m 32s | 18m 49s |
| DiatomSizeReduction | 00m 04s | 00m 03s | 00m 02s |
| ECGFiveDays | 00m 04s | 00m 04s | 00m 04s |
| Haptics | 03m 11s | 03m 12s | 03m 03s |
| ItalyPowerDemand | 00m 25s | 00m 26s | 00m 23s |
| MALLAT | 00m 35s | 00m 27s | 00m 24s |
| MoteStrain | 00m 03s | 00m 03s | 00m 03s |
| TwoLeadECG | 00m 04s | 00m 04s | 00m 04s |
| ArrowHead | 00m 10s | 00m 09s | 00m 11s |
| BirdChicken | 00m 07s | 00m 04s | 00m 04s |
| Herring | 00m 30s | 00m 28s | 00m 30s |
| ProximalPhalanxTW | 03m 39s | 03m 45s | 04m 00s |
| DistalPhalanxOutlineAgeGroup | 01m 42s | 01m 44s | 01m 52s |
| Earthquakes | 02m 02s | 02m 02s | 02m 38s |
| MiddlePhalanxOutlineAgeGroup | 02m 07s | 02m 10s | 02m 46s |
| ShapeletSim | 00m 06s | 00m 04s | 00m 04s |
| Wine | 00m 00s | 00m 00s | 00m 20s |
| Strawberry | 12m 11s | 12m 27s | 12m 09s |

The above steps are repeated for a number of generations $nGen$.

In Figure 5 we show an example of the chromosomes of our algorithm where the dimension of the original time series is 32, whereas the dimension of the reduced space, i.e. $nbp$ is 8.

## 5 Experiments

We conducted three experiments on a large number of time series datasets available at UCR (Chen et al. 2015). This archive contains datasets of different sizes and dimensions and it makes up between 90% and 100% of all publicly available, labeled time series data sets in the world and it represents the interest of the data mining/database community, and not just one group (Ding et al. 2008).

The experiments were conducted on Intel Core 2 Duo CPU with 3G memory using MATLAB.

In our experiments we compared the performance of our method, which uses nature-inspired optimization, against that of traditional time series dimensionality reduction techniques. We implemented our method in three variations that correspond to the three nature-inspired optimization algorithms we presented in Section 4, i.e. GA, DE, and PSO in the case of single-objective optimization, and NSGA-II in the case of multi- objective optimization. As for the traditional dimensionality reduction techniques we compared our method to, we chose PAA and SAX that we presented in Section 4. The main reason why we chose these two dimensionality reduction techniques is, in addition to their popularity, that they both apply compression ratios explicitly.

We chose 30 datasets out of the datasets available in the archive. We meant to choose a variety of datasets to avoid "cherry-picking". The length of the time series on which we conducted our experiments varied between 24 (ItalyPowerDemand) and 1639 (CinC_ECG_torso). The size of the training sets varied between 16 (DiatomSizeReduction) and 390 (Adiac). The size of the testing sets varied between 20 (BirdChicken), (BeetleFly) and 3840 (ChlorineConcentration). The number of classes varied between 2 (Gun-Point), (ECG200), (Coffee), (ECGFiveDays), (ItalyPowerDemand), (MoteStrain), (TwoLeadECG), (BeetleFly), (BirdChicken), (Strawberry), (Herring), (Earthquakes), (ShapeletSim), (Wine), and 37 (Adiac). So as we can see, we tested our method on a diverse range of datasets of different lengths and sizes to avoid getting biased results.

We also have to point out that the results of all our experiments are the average of five runs (except in the case of using PAA and SAX in classification where the results will be the same for all runs so we ran them once only). The control parameters of the three nature-inspired optimization algorithms we applied are shown in Table 1. In order to get unbiased results we chose the same values for the control parameters that the three algorithms GA, DE, and PSO share. These are $popSize$ and $nGen$, and of course $nbp$.

**Table 4** The *k*-means clustering quality of GA, DE, and PSO

| Datasets | k-means Clustering Quality | | | | | | | | | | | | | |
|---|---|---|---|---|---|---|---|---|---|---|---|---|---|---|
| | GA | | | | DE | | | | PSO | | | | PAA | SAX |
| | ρ=1:4 | ρ=1:8 | ρ=1:12 | ρ=1:16 | ρ=1:4 | ρ=1:8 | ρ=1:12 | ρ=1:16 | ρ=1:4 | ρ=1:8 | ρ=1:12 | ρ=1:16 | ρ=1:4 | ρ=1:4 |
| Gun_Point | 0.538 | 0.545 | 0.667 | 0.723 | 0.519 | 0.631 | **0.727** | 0.673 | 0.569 | 0.703 | 0.584 | **0.727** | 0.519 | 0.512 |
| OSULeaf | 0.409 | 0.418 | 0.394 | 0.409 | 0.426 | **0.436** | 0.415 | 0.375 | 0.408 | 0.402 | 0.391 | 0.385 | 0.418 | 0.341 |
| Trace | 0.553 | 0.569 | 0.549 | 0.556 | 0.565 | 0.510 | 0.555 | 0.554 | 0.557 | 0.529 | 0.547 | **0.585** | 0.569 | 0.474 |
| FaceFour | 0.604 | 0.622 | 0.618 | 0.661 | 0.555 | 0.679 | 0.632 | **0.773** | 0.642 | 0.623 | 0.718 | 0.631 | 0.544 | 0.527 |
| ECG200 | 0.705 | 0.770 | 0.750 | 0.793 | 0.752 | 0.780 | **0.801** | 0.773 | 0.743 | 0.746 | 0.778 | 0.795 | 0.620 | 0.584 |
| Adiac | 0.424 | 0.470 | 0.455 | 0.463 | 0.473 | 0.457 | 0.472 | 0.479 | 0.455 | 0.463 | **0.526** | 0.491 | 0.415 | 0.419 |
| FISH | 0.506 | 0.516 | 0.482 | 0.504 | 0.461 | 0.520 | 0.454 | 0.442 | 0.471 | 0.506 | **0.523** | 0.500 | 0.337 | 0.419 |
| Plane | 0.733 | 0.785 | 0.824 | 0.797 | 0.732 | 0.839 | 0.681 | **0.940** | 0.883 | 0.838 | 0.843 | 0.853 | 0.675 | 0.547 |
| Car | 0.611 | 0.583 | 0.560 | 0.575 | 0.513 | 0.576 | 0.595 | **0.628** | 0.510 | 0.559 | 0.591 | 0.593 | 0.528 | 0.448 |
| Beef | 0.458 | 0.444 | 0.481 | 0.481 | 0.446 | 0.450 | 0.481 | 0.481 | 0.481 | **0.511** | **0.511** | 0.494 | 0.444 | 0.385 |
| Coffee | 0.893 | 0.964 | 0.965 | 0.964 | 0.857 | 0.786 | 0.964 | 0.856 | 0.857 | 0.929 | 0.964 | **1.000** | 0.857 | 0.482 |
| CinC_ECG_torso | 0.488 | 0.478 | 0.471 | **0.523** | 0.497 | 0.466 | 0.482 | 0.517 | 0.480 | 0.501 | 0.463 | 0.509 | 0.458 | 0.402 |
| ChlorineConcentration | 0.406 | 0.408 | 0.410 | 0.410 | 0.405 | 0.408 | **0.412** | 0.411 | 0.402 | 0.406 | 0.407 | 0.407 | 0.399 | 0.395 |
| DiatomSizeReduction | 0.787 | 0.786 | 0.964 | **0.976** | 0.961 | 0.792 | 0.764 | 0.765 | 0.765 | 0.767 | 0.959 | 0.763 | 0.821 | 0.482 |
| ECGFiveDays | 0.793 | 0.777 | 0.775 | 0.770 | 0.778 | 0.785 | 0.795 | **0.798** | 0.732 | 0.794 | 0.774 | 0.786 | 0.520 | 0.600 |
| Haptics | 0.331 | 0.340 | 0.356 | 0.340 | 0.359 | 0.336 | 0.358 | 0.349 | 0.351 | 0.352 | **0.379** | 0.371 | 0.325 | 0.292 |
| ItalyPowerDemand | 0.970 | 0.969 | 0.971 | 0.970 | 0.598 | 0.967 | **0.972** | 0.969 | 0.960 | 0.968 | **0.972** | 0.969 | 0.456 | 0.598 |
| MALLAT | 0.871 | 0.914 | **0.938** | 0.848 | 0.870 | 0.852 | 0.867 | 0.860 | 0.747 | 0.870 | 0.930 | 0.864 | 0.873 | 0.612 |
| MoteStrain | 0.806 | 0.792 | 0.826 | 0.707 | **0.879** | 0.858 | 0.870 | 0.832 | 0.800 | 0.877 | 0.819 | 0.832 | 0.805 | 0.363 |
| TwoLeadECG | 0.568 | 0.597 | 0.565 | 0.680 | 0.579 | 0.574 | 0.596 | 0.677 | 0.575 | 0.601 | 0.437 | **0.705** | 0.543 | 0.436 |
| ArrowHead | 0.514 | 0.537 | 0.528 | 0.533 | 0.541 | 0.510 | 0.554 | 0.529 | 0.468 | **0.712** | 0.521 | 0.613 | 0.537 | 0.437 |
| BirdChicken | **0.583** | 0.520 | 0.520 | 0.520 | 0.520 | 0.520 | 0.500 | 0.520 | 0.520 | 0.520 | **0.583** | **0.583** | 0.520 | 0.500 |
| Herring | 0.578 | 0.592 | 0.577 | **0.624** | 0.607 | 0.578 | 0.547 | 0.562 | 0.578 | 0.622 | 0.607 | 0.592 | 0.593 | 0.540 |
| ProximalPhalanxTW | 0.488 | 0.507 | 0.526 | **0.570** | 0.533 | 0.540 | 0.562 | 0.539 | 0.510 | 0.553 | 0.547 | 0.520 | 0.479 | 0.465 |
| DistalPhalanxOutlineAgeGroup | 0.745 | 0.720 | 0.637 | 0.739 | 0.637 | 0.738 | **0.792** | 0.695 | 0.766 | 0.762 | 0.786 | 0.719 | 0.653 | 0.632 |
| Earthquakes | 0.620 | **0.621** | 0.601 | 0.592 | **0.621** | **0.621** | **0.621** | 0.610 | 0.620 | 0.620 | **0.621** | 0.613 | **0.621** | 0.585 |
| MiddlePhalanxOutlineAgeGroup | 0.645 | 0.634 | 0.669 | 0.643 | 0.644 | 0.644 | 0.637 | **0.695** | 0.639 | 0.649 | 0.642 | 0.670 | 0.633 | 0.669 |
| ShapeletSim | 0.544 | 0.572 | **0.575** | 0.566 | 0.567 | 0.538 | 0.533 | 0.539 | 0.521 | 0.561 | 0.515 | 0.550 | 0.521 | 0.502 |
| Wine | 0.610 | 0.590 | 0.550 | 0.600 | 0.550 | 0.667 | **0.721** | 0.682 | 0.600 | 0.610 | 0.600 | 0.600 | 0.533 | 0.491 |
| Strawberry | 0.616 | 0.618 | 0.615 | 0.615 | 0.619 | 0.612 | 0.623 | **0.631** | 0.608 | 0.607 | 0.618 | 0.615 | 0.536 | 0.535 |

Each experiment corresponds to a particular data mining task. The protocol of the three experiments is the same and it consists of two stages: the training stage and the testing stage. In the training stage we perform an optimization process on that dataset to obtain the timestamps which optimize the data mining task in question on that dataset. In the testing stage we strip off of the testing datasets all the data except those that correspond to the optimal timestamps, which we obtained in the training stage, and we perform that data mining task on those reduced datasets.

Applying SAX requires different procedures. In the training stage we search for the alphabet size, among all other values of the alphabet size, that yields the minimum classification error/maximum clustering quality. Then in the testing stage we apply SAX to the corresponding testing dataset using the alphabet size that gives the minimal classification error/maximal clustering quality in the training stage.

The first experiment is a classification task one, i.e. the objective function of the optimization problem is the 1NN classification error, which we aim to minimize. As indicated above, during the training stage we apply the optimization problem on each training dataset and return the timestamps, for that dataset, that minimize the classification error, and then in the testing stage we keep of each time series the data points that correspond to those timestamps, and we classify the resulting data to get the results we show in Table 2. The best result (the minimum classification error – the lower the better) for each dataset is shown in bold printing in yellow-shaded cells. A preliminary comparison shows that nature-inspired optimization algorithms outperform PAA and SAX.

In detail, PSO seems to be the best optimizer for this problem as it gave the best result in 16 datasets out of the 30 datasets tested. This outcome was actually surprising to us because the PSO version we used was a simple one, so may be other versions, particularly adapted to handling integer parameters, could give even better results.

The second best optimizer for this problem is GA, which gave the best result for 13 datasets.

The third best optimizer is DE (which is the optimizer we used in the (Muhammad Fuad 2016)). It turns out this optimizer is not as good as the two other optimizers for this task. DE gave the minimum classification error in 10 datasets.

The performance of the two tested traditional dimensionality reduction techniques is much inferior to that of nature-inspired algorithms. PAA gave the best result for one dataset only (Coffee) and it is in fact the same result obtained when using the three nature-inspired optimization algorithms, whereas SAX gave the best result for two datasets; one (Earthquakes) is even the same result obtained when applying PSO, and the other is (CinC_ECG_torso).

An interesting phenomenon that we notice from the table is that the performance of the three tested nature-inspired algorithms is almost stable even for compression ratios higher than 1:4. In fact, their good performance seems independent of the compression ratio, which is not the case with traditional dimensionality reduction techniques whose performance decreases the higher the compression ratio increases. The explanation that our method, which uses nature-inspired algorithms, gives good results even at higher compression ratios is that our method does not perform any form of direct smoothing, unlike traditional dimensionality reduction techniques which are based on direct smoothing. In other words, the optimization process in our method (whatever the optimizer is) either keeps a data point, which corresponds to an important timestamp, or discards it, whereas traditional dimensionality reduction techniques keep smoothened versions of all the points whether they are important or not. This actually emphasizes the notion of "important points" on which our method is based.

**Table 5** The total score (Ps) of the 1-NN classification error (CE) and *k*-means clustering quality (CQ) of NSGA-II, PAA, and SAX quality of GA, DE, and PSO

| Datasets | NSGA-II | | | | PAA | | | | SAX | | | |
|---|---|---|---|---|---|---|---|---|---|---|---|---|
| | C.E. | C.Q. | Ps C.E. | Ps C.Q. | C.E. | C.Q. | Ps C.E. | Ps C.Q. | C.E. | C.Q. | Ps C.E. | Ps C.Q. |
| Gun_Point | 0.080 | 0.709 | 2 | 2 | 0.093 | 0.519 | 1 | 1 | 0.147 | 0.512 | 0 | 0 |
| OSULeaf | 0.455 | 0.417 | 2 | 1 | 0.488 | 0.418 | 0 | 2 | 0.475 | 0.341 | 1 | 0 |
| Trace | 0.130 | 0.569 | 2 | 2 | 0.250 | 0.569 | 1 | 2 | 0.370 | 0.474 | 0 | 0 |
| FaceFour | 0.170 | 0.659 | 2 | 2 | 0.205 | 0.544 | 1 | 1 | 0.227 | 0.527 | 0 | 0 |
| ECG200 | 0.100 | 0.776 | 2 | 2 | 0.130 | 0.620 | 0 | 1 | 0.120 | 0.584 | 1 | 0 |
| Adiac | 0.358 | 0.475 | 2 | 2 | 0.404 | 0.415 | 1 | 0 | 0.867 | 0.419 | 0 | 1 |
| FISH | 0.200 | 0.509 | 2 | 2 | 0.217 | 0.337 | 1 | 0 | 0.263 | 0.419 | 0 | 1 |
| Plane | 0.010 | 0.833 | 2 | 2 | 0.038 | 0.675 | 0 | 1 | 0.029 | 0.547 | 1 | 0 |
| Car | 0.233 | 0.613 | 2 | 2 | 0.267 | 0.528 | 1 | 1 | 0.267 | 0.448 | 1 | 0 |
| Beef | 0.200 | 0.485 | 2 | 2 | 0.333 | 0.444 | 1 | 1 | 0.433 | 0.385 | 0 | 0 |
| Coffee | 0.000 | 0.964 | 2 | 2 | 0.000 | 0.857 | 2 | 1 | 0.286 | 0.482 | 0 | 0 |
| CinC_ECG_torso | 0.073 | 0.505 | 2 | 2 | 0.104 | 0.458 | 0 | 1 | 0.073 | 0.402 | 2 | 0 |
| ChlorineConcentration | 0.214 | 0.408 | 2 | 2 | 0.390 | 0.399 | 1 | 1 | 0.582 | 0.395 | 0 | 0 |
| DiatomSizeReduction | 0.049 | 0.961 | 2 | 2 | 0.065 | 0.821 | 1 | 1 | 0.082 | 0.482 | 0 | 0 |
| ECGFiveDays | 0.075 | 0.793 | 2 | 2 | 0.146 | 0.520 | 1 | 0 | 0.150 | 0.600 | 0 | 1 |
| Haptics | 0.581 | 0.363 | 2 | 2 | 0.643 | 0.325 | 1 | 1 | 0.643 | 0.292 | 1 | 0 |
| ItalyPowerDemand | 0.034 | 0.970 | 2 | 2 | 0.068 | 0.456 | 1 | 0 | 0.192 | 0.510 | 0 | 1 |
| MALLAT | 0.080 | 0.870 | 2 | 1 | 0.089 | 0.873 | 1 | 2 | 0.143 | 0.612 | 0 | 0 |
| MoteStrain | 0.144 | 0.854 | 2 | 2 | 0.190 | 0.805 | 1 | 1 | 0.212 | 0.707 | 0 | 0 |
| TwoLeadECG | 0.231 | 0.621 | 2 | 2 | 0.283 | 0.543 | 1 | 1 | 0.309 | 0.436 | 0 | 0 |
| ArrowHead | 0.189 | 0.613 | 2 | 2 | 0.206 | 0.537 | 1 | 1 | 0.246 | 0.437 | 0 | 0 |
| BirdChicken | 0.250 | 0.520 | 2 | 2 | 0.450 | 0.520 | 0 | 2 | 0.350 | 0.500 | 1 | 0 |
| Herring | 0.391 | 0.606 | 2 | 2 | 0.484 | 0.593 | 0 | 1 | 0.406 | 0.540 | 1 | 0 |
| ProximalPhalanxTW | 0.262 | 0.545 | 2 | 2 | 0.280 | 0.479 | 1 | 1 | 0.370 | 0.465 | 0 | 0 |
| DistalPhalanxOutlineAgeGroup | 0.205 | 0.742 | 2 | 2 | 0.235 | 0.653 | 1 | 1 | 0.267 | 0.632 | 0 | 0 |
| Earthquakes | 0.180 | 0.621 | 2 | 2 | 0.311 | 0.621 | 0 | 2 | 0.180 | 0.585 | 2 | 0 |
| MiddlePhalanxOutlineAgeGroup | 0.247 | 0.656 | 2 | 2 | 0.277 | 0.633 | 0 | 1 | 0.242 | 0.618 | 1 | 0 |
| ShapeletSim | 0.417 | 0.552 | 2 | 2 | 0.472 | 0.521 | 0 | 1 | 0.428 | 0.521 | 1 | 1 |
| Wine | 0.315 | 0.667 | 2 | 2 | 0.370 | 0.533 | 1 | 1 | 0.500 | 0.491 | 0 | 0 |
| Strawberry | 0.044 | 0.612 | 2 | 2 | 0.062 | 0.536 | 1 | 1 | 0.328 | 0.535 | 0 | 0 |
| | | | **118** | | | | **53** | | | | **18** | |

It is important to mention here that while processing time can be one of the main downsides of applying nature-inspired optimization algorithms to data mining, our method is in fact quite fast for the three optimizers we used. In Table 3 we show the training time of the classification task for the experiments we presented in Table 2. The results shown are those of compression ratio 1:4 (the most time-consuming compression ratio). As we can see from the table, the training time takes between few seconds to few minutes. In fact, for some datasets the training time of 100 generations was almost instantaneous and took less than a second (these are the datasets with running time 00m 00s in the table). As a matter of fact, the longest training time did not exceed 19m 32s (ChlorineConcentration with DE). We also have to take into account that our codes were not optimized for speed. Besides, we are using MATLAB, which is not very fast.

The second experiment was on clustering, i.e. the objective function of the optimization problem is the $k$-means clustering quality, which we aim to maximize (we however, process this as a minimization optimization problem, which is a convention in optimization). During the training stage, we apply the optimization problem on each training dataset to obtain the timestamps that optimize the clustering quality, and then in the testing stage we cluster the testing datasets based on the data that correspond to those timestamps. In Table 4 we present the results we obtained. The best result (the maximum clustering quality – the higher the better) for each dataset is shown in bold printing in yellow-shaded cells. As we can see from the table, the best clustering results are obtained by using DE as this optimization algorithm gives the best result in 15 datasets. The second best optimization algorithm is PSO as it gives the best result in 12 datasets. GA comes third as it gives the best results in 8 datasets. The performance of PAA and SAX is quite inferior to the three nature-inspired methods. In fact, PAA gives the best clustering quality in one dataset only (Earthquakes), which is the same result obtained by using the three nature-inspired optimization methods.

The third experiment is a multi-objective optimization one, i.e., we have two objective functions: classification error and clustering quality. The optimizer that we choose is NSGA-II that we presented in Section 4. Performance evaluation in this case is not trivial because a method A can outperform another method, B, on one task whereas method B can outperform method A on the other task. In order to evaluate the performance of the three methods, NSGA-II, PAA and SAX, we adopted a criterion which is similar to the one we applied in (Muhammad Fuad 2015) that we explain in the following: for each dataset, and for each task, we give the method (NSGA-II, PAA or SAX) that yields the best result for that task 2 points, then 1 point for the second best and 0 for the last one. In case where two methods yield the same result we give them the same points and we skip the points for the following rank. So for the two tasks together each method can obtain a maximum of 4 points, then we take the sum of all the points each method obtained on all the datasets tested. The final results are shown in Table 5. As we can see from the results, for the two tasks together, $k$-means clustering and 1NN classification, the best performance is that of NSGA-II as this method obtains 118 points out of 120 possible points. This is much better than the performance of the second best method, PAA, which obtains 53 points. SAX comes last with 18 points only.

## 6 Conclusion

In this work we presented a new method for time series dimensionality reduction through an optimization process. Unlike traditional time series representation methods based on landmark points or on data transformations, our method combines the advantages of these two categories by handling the dimensionality reduction process as an optimization problem whose outcome is the timestamps of all the time series in the dataset that yield the optimal performance of a time series mining task on that dataset. This approach has several advantages; the first is that the Euclidean distance defined on the low-dimension representation of the data is a lower bound of the Euclidean distance defined on the raw data, which allows the application of the GEMINI algorithm. The other advantage is that we can have a fully customized low-dimension representation of the data as the user can define beforehand the exact dimension of the reduced space instead of having this dimension determined indirectly by means of a compression ratio. Another advantage is the high performance of this method in performing time series tasks.

We implemented our method using three widely known nature- inspired optimizations algorithms: GA, DE, and PSO, in addition to NSGA-II in the case of multi-objective optimization. However, other nature-inspired optimization algorithms can also be used for this purpose. We showed how the optimization process itself can be performed quite fast.

We chose to apply our method to the two main time series mining tasks; classification and clustering, but our method can be applied to other tasks that require dimensionality reduction, whether on individual time series mining tasks, using single-objective optimization, or on several tasks simultaneously, using multi-objective optimization.

In the future, we hope to extend our method to other data types. Another direction of future research is to perform the optimization process in a more data-oriented manner.